\documentclass[runningheads]{llncs}
\usepackage{graphicx}
\usepackage{subfigure}
\usepackage{amsmath}
\usepackage{amsfonts}
\usepackage{graphicx}
\usepackage{color} 
\usepackage{hyperref}
\newcommand{\R}{\mathbb{R}}
\newcommand{\FN}{\text{FN}}

\newcommand{\FP}{\text{FP}}
\newcommand{\TP}{\text{TP}}
\DeclareMathOperator{\argmax}{argmax}
\usepackage{enumitem}

\newcommand\blfootnote[1]{%
  \begingroup
  \renewcommand\thefootnote{}\footnote{#1}%
  \addtocounter{footnote}{-1}%
  \endgroup
}

\begin{document}
\title{Deep Learning Based Rib Centerline Extraction and Labeling} 
\author{Matthias Lenga, Tobias Klinder, Christian B\"{u}rger, Jens von Berg,\\ Astrid Franz, Cristian Lorenz}
\authorrunning{ }
\institute{Philips Research Europe, Hamburg, Germany}
\maketitle              
\begin{abstract} Automated extraction and labeling of rib centerlines is a typically needed prerequisite for more advanced assisted reading tools that help the radiologist to efficiently inspect all 24 ribs in a CT volume. In this paper, we combine a deep learning-based rib detection with a dedicated centerline extraction algorithm applied to the detection result for the purpose of fast, robust and accurate rib centerline extraction and labeling from CT volumes. More specifically, we first apply a fully convolutional neural network (FCNN) to generate a probability map for detecting the \emph{first rib} pair, the \emph{twelfth rib} pair, and the collection of all \emph{intermediate ribs}. In a second stage, a newly designed centerline extraction algorithm is applied to this  multi-label probability map. Finally, the distinct detection of first and twelfth rib separately, allows to derive individual rib labels by simple sorting and counting the detected centerlines. We applied our method to CT volumes from 116 patients which included a variety of different challenges and achieved a centerline accuracy of 0.787 mm with respect to manual centerline annotations.

\blfootnote{This is a preprint version of \cite{ML19}.
The final authenticated publication is available 
online at \doi{10.1007/978-3-030-11166-3\_9}}

\end{abstract}

\section{Introduction}\label{sec:intro}
The reading of the ribs from 3D CT scans is a typical task in radiology, e.g., to find bone lesions or identify fractures. During reading, each of the 24 ribs needs to be followed individually while scrolling through the slices. As a result, this task is time-consuming and rib abnormalities are likely to be overlooked. 

In order to assist reading, efficient visualization schemes or methods for navigation support are required. These applications are typically based on the rib centerlines, cf. \cite{Gomez18,Wu08}. Despite their generally high contrast, automated extraction of the rib centerlines from CT is challenging. For example, image noise and artifacts impede the extraction, but also other bony structures in close vicinity (most prominently the vertebra), as well as severe pathologies. Finally, anatomical labeling of the extracted centerlines (i.e. knowing which one for example is the ``7th right rib'') is usually desirable. From an algorithmic perspective, this task is trivial if all 24 ribs are correctly extracted, as simply counting left and right ribs from cranial to caudal would be sufficient. Obviously, this task becomes significantly more challenging once the rib cage is only partially imaged or once a rib is missing (e.g., due to pathologies or missed detection in a previous step).

A wide range of different approaches has been proposed in the past for rib centerline extraction partially also including their labeling. Tracing based approaches, as in \cite{Shen04,Lee10} aim at iteratively following the ribs. As such approaches rely on an initial seed point detection per rib, an entire rib is easily missed once a corresponding seed point was not detected. Alternatively, potential rib candidates can be first detected in the entire volume which then need to be grouped to obtain ribs, as for example done in \cite{Staal06}. However, the removal of other falsely detected structures remains a crucial task. Attempts have been made to additionally integrate prior knowledge by means of geometrical rib cage centerline models, cf. \cite{TK08,Wu08}. Nevertheless, such approaches may struggle with deviations from the model in terms of pathologies.

In this paper, we propose a two-stage approach combining deep learning and classic image processing techniques to overcome several of the limitations listed above. Rib probability maps are calculated at first using a fully convolutional neural network, see Subsection \ref{subsec:approach}, and then the centerlines are reconstructed using a specifically designed centerline extraction algorithm as described in Subsection \ref{subsec:walker}. In particular, three distinct rib probability maps are calculated (\emph{first rib}, \emph{twelfth rib} or \emph{intermediate rib}). By knowing the first and/or twelfth rib, labeling can be solved easily by iterative counting. This scheme also works in case of partial rib cages (for example if only the upper or lower part is shown). Evaluation is carried out on a representative number of 116 cases.

\section{Methods}
\subsection{Data}\label{subsec:data}
Our data set consists in total of 116 image volumes containing 62 thorax as well as 54 full body CT scans. The data includes a wide range of typical challenges, such as variation in the field of view leading to partly visible or missing ribs (3 patients with first rib missing, 38 patients with at least partially missing twelfth rib), various types of rib fractures, spine scoliosis (14 patients) strong contrast-uptake around the first rib (33 patients), implants in other bony structures (7 around the sternum, 2 around the spine, and 2 around the femur/humerus), several different devices with similar intensity to the ribs such as catheters or cables (57 patients). 

In each image, we annotated rib centerlines by manually placing spline control points. The rib centerlines were then obtained using cubic spline interpolation. For each image volume, we generated a label mask by dilating the corresponding centerlines with a radius of 3.0 mm. Four different labels are assigned to the classes \emph{background}, \emph{first rib}, \emph{twelfth rib} and \emph{intermediate rib}.

\subsection{Multi-Label Rib Probability Map Generation}\label{subsec:approach}
For rib detection, we first apply a fully convolutional neural network (FCNN) in order to generate probability maps which are subsequently fed into the tracing algorithm described in Subsection \ref{subsec:walker}. 
More specifically, we formulate our task as a 4-class problem, where the network yields for each voxel $v_{ijk}$ of the volume a 4-dimensional vector $p_{ijk} \in [0,1]^4$. The components $p_{ijk, 0}, p_{ijk, 1}, p_{ijk, 2}, p_{ijk, 3}$ can be interpreted as probabilities that the associated voxel belongs to the classes \emph{background}, \emph{first rib (pair)}, \emph{twelfth rib (pair)} or \emph{intermediate rib (pairs)}, respectively. Distinct classes for the first and the twelfth rib were introduced to deal with differences in anatomy (especially for the first rib) while significantly simplifying the following labelling procedure. By using the relative to location of the intermediate ribs to the first and twelfth rib, labelling of the ribs can be achieved efficiently. Moreover, knowing the potential location of first or twelfth rib enables labelling even in cases of partial rib cages. Details are provided in Subsection \ref{subsec:walker} below.

We favored the parsimonious 4-class learning task over training a neural network for detecting each individual rib, resulting in a 25-class (24 ribs plus background) classification problem, due to several reasons: i) The 4-class network in combination with our iterative tracing approach seems sufficient for solving the problem at hand, ii) due to the similar appearance of intermediate ribs, we do not expect the 25-class network to be able to identify internal ribs reliably, iii) the 25-class approach would cause a higher memory footprint and runtime during training and inference. 

\begin{figure}
	\centering
	\includegraphics[width=0.75\linewidth]{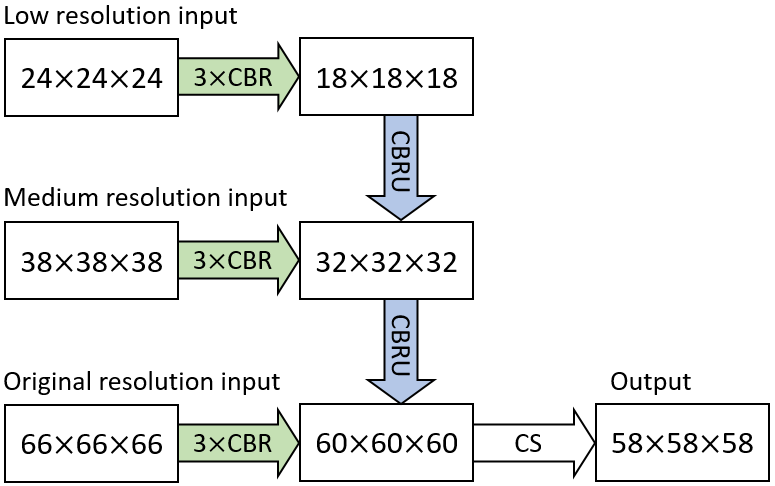}
	\caption{Foveal architecture with 3 resolution levels. The feature extraction pathways (green), consisting of 3 CBR blocks, are integrated using CBRU blocks (blue). The final CS block consists of a $3\times 3 \times 3$ valid convolution and a soft-max layer. }
 \label{fig:FNet}
\end{figure}

As network architecture, we chose the Foveal network described in \cite{BS18}. Basically, the network is composed of two different types of layer modules, CBR and CBRU blocks, see Figure \ref{fig:FNet}. A CBR block  consists of a $3\times 3 \times 3$ valid convolution (C) followed by batch normalization (B) and a rectified linear unit activation (R). A CBRU block is a CBR block followed by an average unpooling layer (U).
Since we favor fast network execution times and a moderate GPU memory consumption, we decided to use three resolution layers $L_\text{H}, L_\text{M}, L_\text{L}$, each composed of three CBR blocks. Differently sized image patches with different isotropic voxel spacings are fed into the layers as input, see Table \ref{Tab:Resolutions}.
The low and medium resolution pathways $L_\text{L}, L_\text{M}$ are integrated into the high resolution layer $L_\text{H}$ using CBRU blocks. Implementation details and further remarks concerning the architecture performance can be found in \cite{BS18}.

\begin{table}
\centering
\begin{tabular}{c|c|c}
& input patch size (voxel) & patch voxel spacing (mm) \\
\hline
$L_\text{H}$ original resolution & $66 \times 66 \times 66$ & $1.5 \times 1.5 \times 1.5$ \\
$L_\text{M}$ medium resolution & $38 \times 38 \times 38$ & $3.0 \times 3.0 \times 3.0$ \\
$L_\text{L}$ low resolution & $24 \times 24 \times 24$ & $6.0 \times 6.0 \times 6.0$ \\
\multicolumn{3}{c}{~~}\\
\end{tabular}
\caption{~Input configuration of the network layers.}\label{Tab:Resolutions}
\end{table}

As preprocessing, the CT images are resampled to an isotropic spacing of 1.5 mm using linear interpolation and normalized to zero mean and unit standard deviation. 
The network was trained by minimizing the cross entropy on mini-batches containing 8 patches (each at three different resolutions) drawn from 8 randomly selected images. In order to compensate for the class imbalance between background and rib voxels, we used the following randomized sampling strategy: 10\% of the patch centers were sampled from the bounding box of the first rib pair, 10\% from the bounding box of the twelfth rib pair and 30\% from the bounding box of the intermediate ribs. The remaining 50\% patch centers were uniformly sampled from the entire volume. As an update rule, we chose AdaDelta \cite{ZE12} in combination with a learning rate schedule. For data augmentation, the patches were randomly scaled and rotated around all three axes. The neural network was implemented with CNTK version 2.2 and trained for 2000 epochs on a GeForce GTX 1080. The network training could be completed within a few hours and network inference times were ranging from approximately 5 to 20 seconds, depending on the size of the input CT volume.

\subsection{Centerline Extraction and Labeling}\label{subsec:walker}

In order to robustly obtain rib centerlines, we designed an algorithm that specifically incorporates the available information from the multi-label probability map. It basically consists of four distinct steps:
\begin{enumerate}
	\item Determination of a rib cage bounding box. 
	\item Detection of an initial left and right rib.
	\item Tracing of the detected ribs and detecting neighboring ribs iteratively upwards and downwards of the traced rib.
	\item Rib labeling.
\end{enumerate}
Steps 1 to 3 are performed on the \emph{combined probability map}, adding the results of the three non-background classes and limiting the sum to a total probability of 1.0, i.e. to each voxel $v_{ijk}$ we assign the value $q_{ijk} := \min\{ p_{ijk, 1} + p_{ijk, 2} + p_{ijk, 3}, 1 \}$.\\

\begin{figure}
	\centering
	\subfigure[]{\includegraphics[height=65mm]{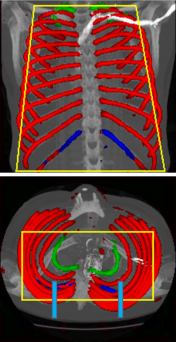}}
	\subfigure[]{\includegraphics[height=65mm]{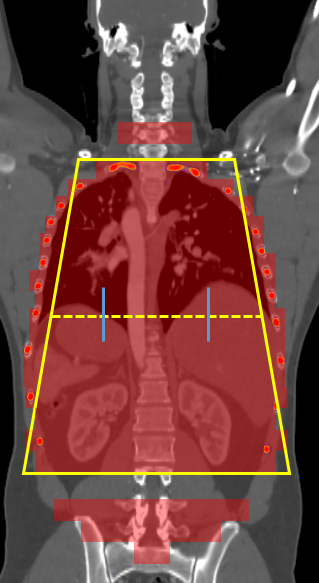}}
	\subfigure[]{\includegraphics[height=65mm]{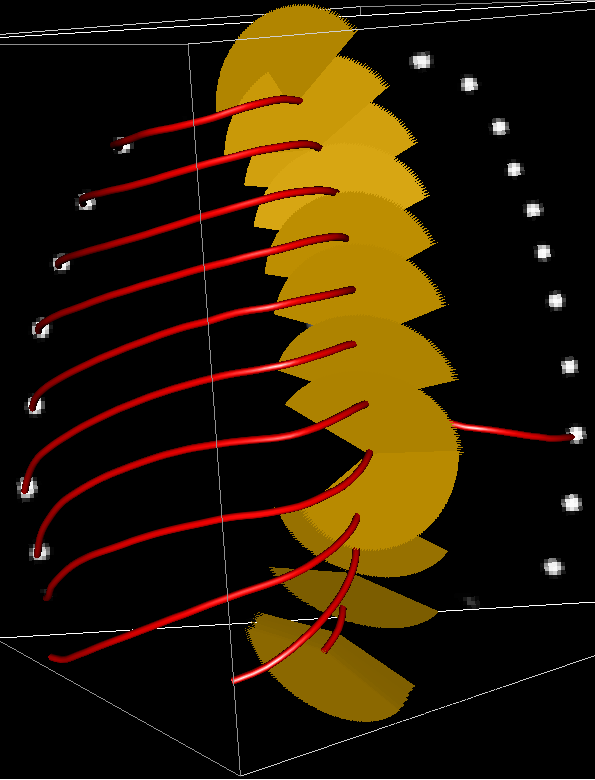}}
	\caption{(a) Neural network output (green: first rib; red: intermediate rib; blue: twelfth rib) and approximate 3D bounding box of the rib cage (yellow) in coronal (top) and axial view (bottom). The lower image depicts in light blue the two search regions for rib detection. (b) Schematic representation of the vertical stack of 2D bounding boxes (red) in coronal view and the approximate 3D bounding box of the rib cage resulting from the largest connected stack in vertical direction by linear regression (yellow). The dashed yellow line marks the box section at medium axial level. The two search regions used for initial rib detection are depicted in light blue.
(c) Traced ribs (red) are shown on top of a sagittal cross-section of the probability map. The fan-like search regions for neighboring ribs are depicted in yellow.
 \label{fig:tracing}}
\end{figure}

\noindent\textbf{Step 1: Bounding Box Detection}\\
Generally, the given CT volume is assumed to cover at least a large portion of the rib cage, but may extend beyond it.
Therefore, we first determine a search region in order to identify the visible ribs. Based on the axial image slices, a 2D bounding rectangle is computed using a probability threshold of 0.5 on the combined probability map. To suppress spurious responses, we require a minimal 2D box size of 30\,mm$\times$10\,mm to be a \emph{valid} bounding box. 
From the set of valid 2D bounding boxes, a 3D bounding box is calculated from the largest connected stack in vertical direction. The 3D bounding box is strictly speaking not a box, but has inclined faces. Each of the 4 faces results from a linear regression of the slice wise determined 4 border positions, having the advantage of being robust against outliers and being able to represent to some extent the narrowing of the rib cage from abdomen to shoulders (see Figure \ref{fig:tracing} a,b).\\

\noindent\textbf{Step 2: Initial Rib Detection}\\
From the approximate rib cage bounding box obtained in Step 1, we derive an initial cross-sectional search window to detect the ribs. For that purpose, anchor point $a_l, a_r$ are chosen at 25\% and 75\% of the left-to-right extension of the box section at medium axial level.
Then sagittal 2D search regions centered at $a_l$ and $a_r$ of spacial extension 100\,mm$ \times$ 100\,mm are defined (see Figure \ref{fig:tracing} a,b).
In each of these regions an initialization point exceeding a probability of 0.5 is determined. We remark that this point may be located at the rib border. To locate the rib center, we sample the probability values in a spherical region of 15\,mm diameter around the initialization point. 
Next, the probability weighted center of mass $c_0\in \R^3$ and the probability weighted covariance matrix $\Sigma_0 \in \R^{3\times 3}$ of the voxel coordinates are calculated.
Finally, we use $c_0$ as rib center estimate and the eigenvector $t_0\in\R^3$ corresponding to the largest eigenvalue of $\Sigma_0$ as estimation of the tangential direction. The position $c_0$ is added to the list of rib center line control points.\\

\noindent\textbf{Step 3: Rib Tracing and Detection of Neighboring Ribs}\\
Based on the initial rib detection result from Step 2, the rib response in the probability map is traced in an iterated scheme ($i = 0,1,2...$) consisting of
the following three actions:
\begin{enumerate}[label=\alph*)]
\item Starting from $c_i$ move in tangential direction $t_i$ until a voxel with combined probability value $q_{ijk} < 0.5$ is encountered or a maximal moving distance of 7.5 mm is reached.
\item Calculate the weighted mean vector $c_{i+1}\in\R^3$ in a spherical region around the current position. Add $c_{i+1}$ to the list of rib center line control points and move to $c_{i+1}$.
\item Calculate the probability weighted covariance matrix $\Sigma_{i+1} \in \R^{3\times 3}$ in a spherical region around $c_{i+1}$ and compute the tangential direction $t_{i+1}\in \R^3$, see Step 2.
\end{enumerate}

This scheme is iterated until the moving distance 
in the current iteration falls below a predefined threshold of 3.0 mm.  
In that case, a forward-looking mechanism is triggered which aims at bridging local drop-outs of the probability response. More precisely, the algorithm searches for a voxel with a combined probability value exceeding 0.5 within a cone-shaped region. This voxel then serves as continuation point for the iterative tracing procedure described above.

\begin{figure}
	\centering
	\includegraphics[width=0.75\linewidth]{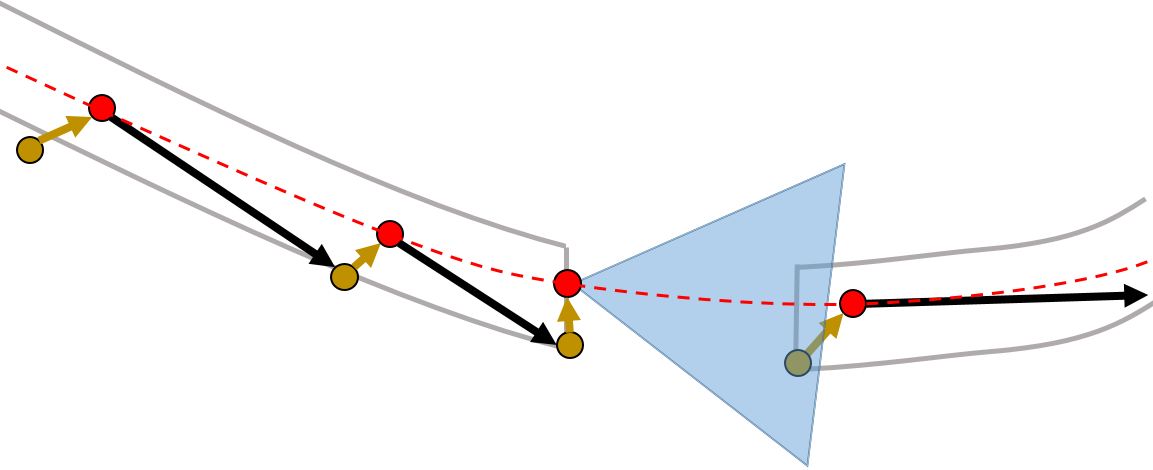}
	\caption{Schematic representation of the iterative tracing algorithm. Each red point corresponds to a probability weighted mean vector $c_i$ in the spherical region around the associated preceding position which depicted by a yellow point connected by a yellow arrow (see Step 3b). The black arrows correspond to a movement in tangential direction $t_i$ (see Step 3a). The blue triangle depicts the cone-shaped search region used by the forward-looking mechanism. The rib center line resulting from a spline interpolation of the control points $c_i$ is depicted by the dashed red line.}
 \label{fig:iterative_scheme}
\end{figure}

Tracing from the initial starting point $c_0$ is performed in both possible directions and results are finally concatenated which yields a centerline of the full rib represented by  the point sequence $\{c_0, c_1, ...\}$.
After the tracing of one rib is completed, the resulting centerline is inserted into 
the list of rib centerlines $L$ which is ordered in vertical direction from feet to head.

This collection is extended in a step wise fashion by detecting
adjacent so far untraced ribs using fan-like 2D search regions anchored at the lowest and highest rib contained in $L$ (see Figure \ref{fig:tracing}b).
 
The initial location of the search fan is 10\,mm distal from the rib starting point at the spine. The rib tangential vector at this point is used as normal vector of the fan plane. 
The fan opening direction withing this plane is steered by the intersection point of the previous rib with the fan plane. If only one traced rib is available yet, the fan is simply pointing upward or downward. 
If a neighboring rib could be found within the fan, the iterative scheme described above is applied to trace the rib. If not, the search fan is moved along the rib in 10\,mm steps towards the distal end.\\

\noindent\textbf{Step 4: Rib Labeling}\\
After extraction of the centerlines, the average probability for all three non-background classes is calculated for each found rib. In the optimal case, 12 rib pairs have been found and the first and twelfth rib have an average probability along their centerlines above 0.5 for their respective class. In this case, the intermediate ribs are labeled according to their position in the list $L$. 
In case that less than 12 ribs were traced, the labeling is still possible if either the  first or twelfth rib can be identified. 
Labeling is not possible if both first and twelfth rib cannot be identified and less then 10 ribs were traced.

\section{Results}
Our pipeline was evaluated using 4-fold cross validation (CV). More precisely,
the dataset was randomly shuffled and partitioned into 4 equally sized subsamples each containing 29 images. We trained 4 different networks by using in each fold three subsamples as training data while retaining a single subsample as validation data for testing. In this way, it is ensure that each data set was contained once in a testing subsample and as a result one probability map was obtained per case.

\subsection{Multi-Label Network}
For the evaluation of a probability map $p_{ijk} = (p_{ijk, 0}, p_{ijk, 1}, p_{ijk, 2}, p_{ijk, 3})$ generated by the neural network, we assigned to each voxel $v_{ijk}$ a predicted class label $L^\text{pred}_{ijk}$ based on its maximal class response, i.e. $$L^\text{pred}_{ijk} = \argmax_{c=0,1,2,3}\, p_{ijk, c}.$$
Following the naming convention from Subsection \ref{subsec:approach}, the labels $0,1,2,3$ correspond to the classes \emph{background}, \emph{first rib}, \emph{twelfth rib} and \emph{intermediate rib}, respectively. Comparing the predicted class labels with the corresponding ground truth labels $L^\text{GT}_{ijk}$, yields for each class the number of true positives (TP), false positives (FP), and false negatives (FN), i.e.
\begin{align*}
\text{TP}_C &=  \vert\{ ijk ~ : ~ L^\text{GT}_{ijk} = C \text{ and } L^\text{pred}_{ijk} = C \}\vert\\
\text{FP}_C &=  \vert\{ ijk ~ : ~ L^\text{GT}_{ijk} \not= C \text{ and } L^\text{pred}_{ijk} = C \}\vert\\
\text{FN}_C &=  \vert\{ ijk ~ : ~ L^\text{GT}_{ijk} = C \text{ and } L^\text{pred}_{ijk} \not= C \}\vert
\end{align*}
where $C\in\{0,1,2,3\}$ denotes the class under consideration. Henceforth, we will omit the class index $C$ in order to simplify the notation. Based on these quantities we compute for each class sensitivity, precision and Dice as follows:
\begin{equation}\label{eq:measures}
\begin{aligned}
\text{sensitivity}&= \frac{\TP}{\TP+\FN}\\
\text{precision}&= \frac{\TP}{\TP+\FP}\\
\text{Dice}&=  \frac{2\TP}{2\TP+\FP+\FN}
\end{aligned}
\end{equation}
Table \ref{table:measures1} displays the statistics of the aforementioned measures calculated on the label images 
contained in the test sets from the 4-fold CV. For the class labels \emph{first rib} and \emph{intermediate rib}
all 116 images were considered. For the class label \emph{twelfth rib} we excluded 21 images from our evaluation which did not contain any part of the twelfth rib pair. 

In order to analyze the overall rib detection rate irrespective of the specific rib class, we assigned a single label to each non-background voxel. Based on these combined masks, we again calculated the statistical measures from Equation \ref{eq:measures} on all 116 images.
The obtained results are summarized in Table \ref{table:measures1} as class label \emph{rib}.
\begin{center}
\begin{table}
\begin{minipage}[b]{0.2\linewidth}
\centering
\begin{tabular}{c|c|c|c}
	\emph{first rib} 		& \bf{sens.} 	&	\bf{prec.}	& \bf{Dice} \\
\hline
mean	& 0.65	& 0.70	& 0.67\\
std.	& 0.13	& 0.12	& 0.12\\
25\% qrt.	& 0.58	& 0.66	& 0.62\\
median	& 0.66	& 0.73	& 0.70\\
75\% qrt.	& 0.74	& 0.78	& 0.74\\
\multicolumn{4}{c}{~~}\\

\emph{intermediate rib} 		& \bf{sens.} 	&	\bf{prec.}	& \bf{Dice} \\
\hline
mean		&0.81			&0.87			&0.84\\
std.		&0.07			&0.04			&0.05\\
25\% qrt.	&0.79			&0.84			&0.82\\
median	&0.82			&0.87			&0.84\\
75\% qrt.	&0.85			&0.90			&0.87\\
\multicolumn{4}{c}{~~}\\
\end{tabular}	
\end{minipage}\hfill
\begin{minipage}[b]{0.4\linewidth}
\centering
\begin{tabular}{c|c|c|c}
\emph{twelfth rib} 		& \bf{sens.} 	&	\bf{prec.}	& \bf{Dice} \\
\hline
mean	&0.60	&0.63	&0.59\\
std.	&0.22	&0.23	&0.20\\
25\% qrt.	&0.49	&0.54	&0.47\\
median	&0.66	&0.71	&0.64\\
75\% qrt.	&0.77	&0.81	&0.74\\
\multicolumn{4}{c}{~~}\\
\emph{rib} 		& \bf{sens.} 	&	\bf{prec.}	& \bf{Dice} \\
\hline
mean	&0.81	&0.87	&0.84\\
std.	&0.07	&0.04	&0.05\\
25\% qrt.	&0.79	&0.84	&0.82\\
median	&0.82	&0.87	&0.84\\
75\% qrt.	&0.84	&0.90	&0.86\\
\multicolumn{4}{c}{~~}\\
\end{tabular}
\end{minipage}\hfill
\caption{Mean, standard deviation, 25\% quartile, median and 75\% quartile of the statistical measures for the predicted class labels \emph{first rib}, \emph{intermediate rib}, \emph{twelfth rib} and the combined class \emph{rib}.}\label{table:measures1}
\end{table}
\end{center}
\vspace*{-1cm}
As can be seen from Table \ref{table:measures1}, we obtain overall good performance for the overall rib detection captured for example with an mean Dice of 0.84. Let us remark that for thin objects, such as the dilated rib centerlines, the Dice score constitutes a rather sensitive measure. The results indicate that detecting the first and twelfth rib pairs is more difficult for our network. While extraction of the first rib is more challenging due to, e.g., higher noise in the upper thorax or other bony structures in close vicinity (clavicle, shoulder blades, vertebrae), the twelfth rib can be extremely short and is easily confused by the neighboring ribs. For further illustration, Figure \ref{MIPS_difficult3} shows the results on selected representative cases. Generally, the ribs are well detected without major false responses in other structures - despite all the different challenges present in the data. The color coding highlighting of the multi-label detection reveals that first and twelfth are mostly correctly detected. In few cases the network wrongly generated strong responses of the classes \emph{first rib} or \emph{last rib} for voxels belonging to the second or eleventh rib pair.

\subsection{Rib centerlines}
For the evaluation of the final centerlines, both ground truth lines and automatically determined centerlines were resampled to 1.0 mm uniform point distance. A \emph{true positive distance} of $\delta$ = 5.0 mm was chosen such that, if for a ground truth point (GTP) no result point within $\delta$ was found, the GTP was counted as false negative (FN). Result points having a corresponding GTP within $\delta$ were counted as true positive (TP), all other as false positive (FP). From the TP, FP, and FN values we calculated sensitivity, precision and Dice using Equation \eqref{eq:measures}.

Table \ref{tbl_result_ctr_line_labeled} summarizes our results from the 4-fold cross-validation. The point wise responses (TP, FP, FN) are averaged up over all cases. The evaluation measures are finally reported on a per rib basis, as well as for all ribs. The Euclidean distance (dist.) is measured as point-to-line distance between result point and ground truth line. 
Moreover, Table \ref{tbl_result_ctr_line_casewise} contains the percentage of cases with missed labeled ribs. Here, a rib is counted as missed, if less than half of the ground truth rib centerline could be detected. A detected rib centerline point counts only as true positive if the correct label was determined.

\begin{figure}
\centering
\subfigure[]{\includegraphics[height=60mm]{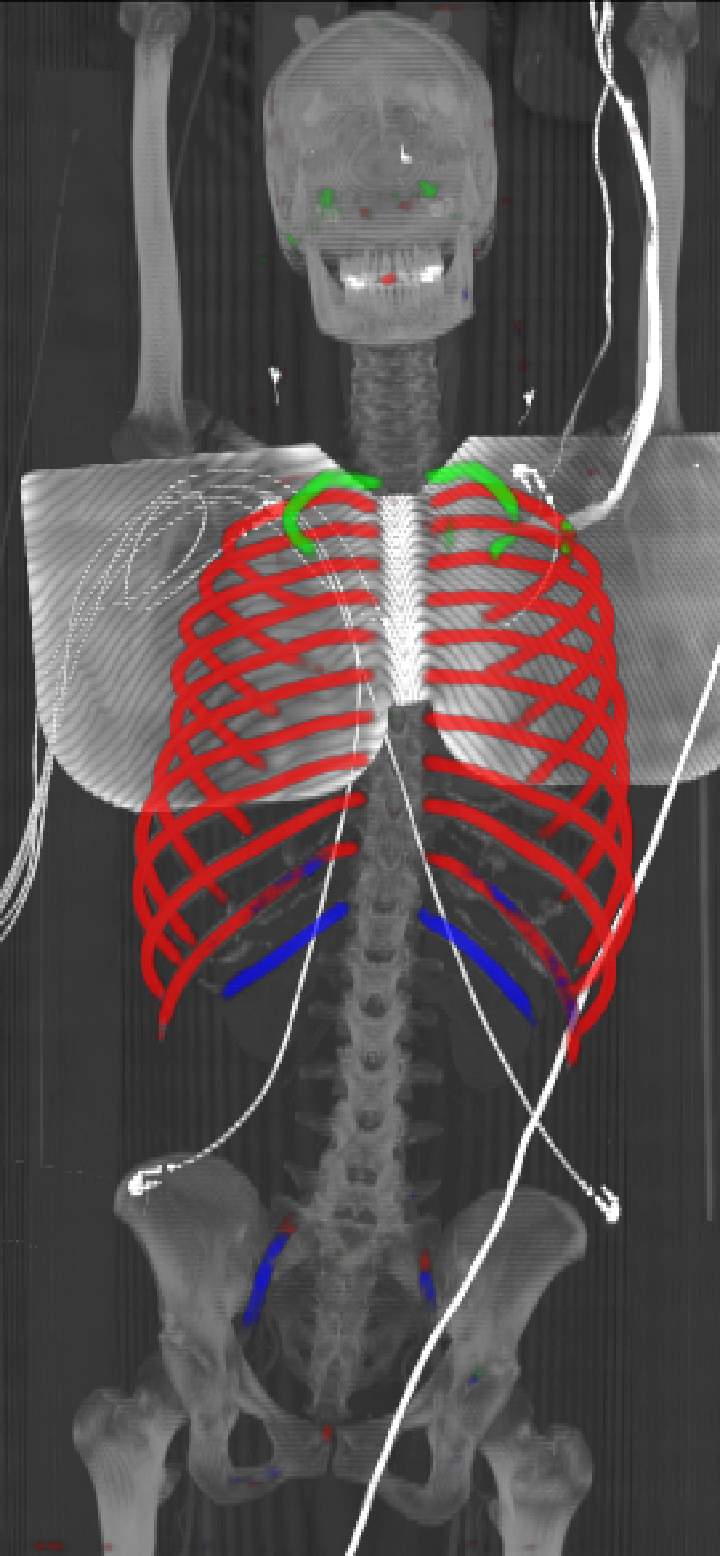}}
\subfigure[]{\includegraphics[height=60mm]{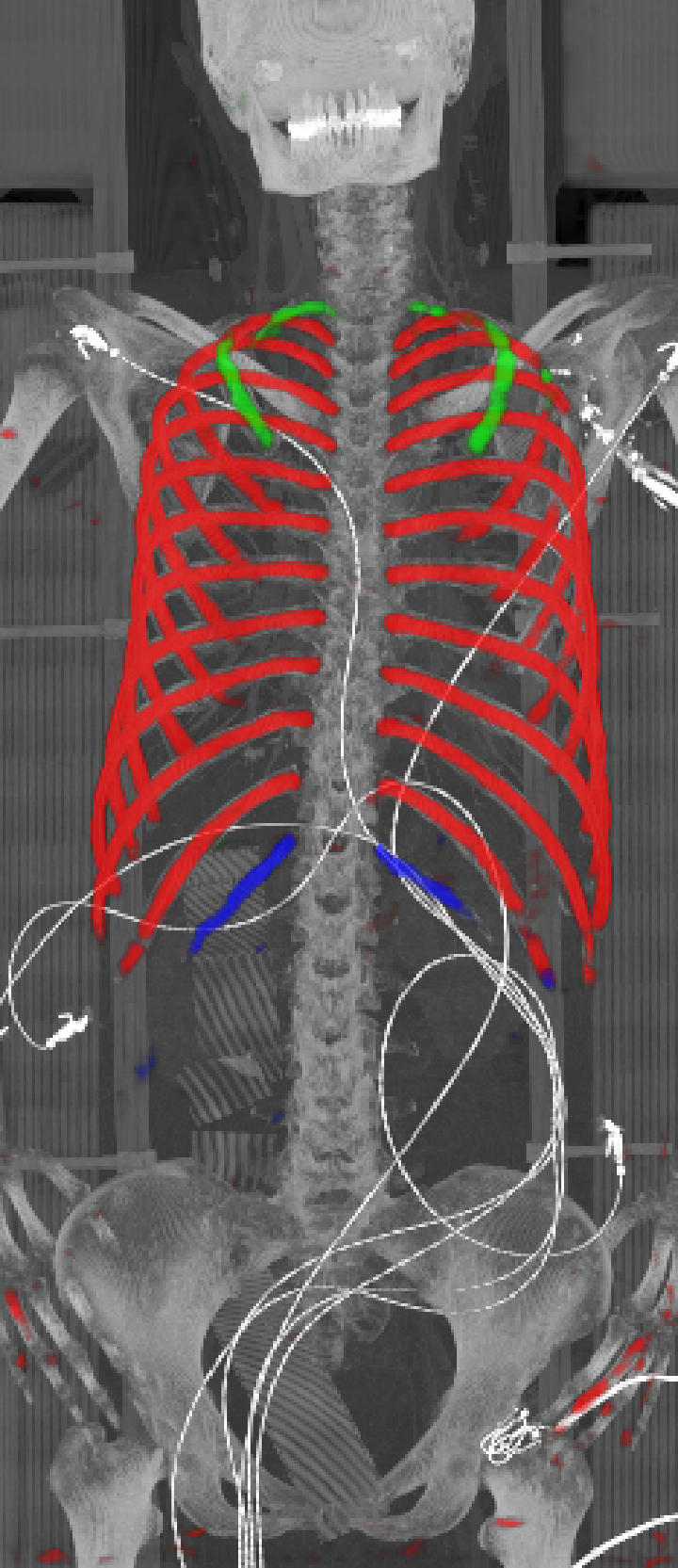}}
\subfigure[]{\includegraphics[height=60mm]{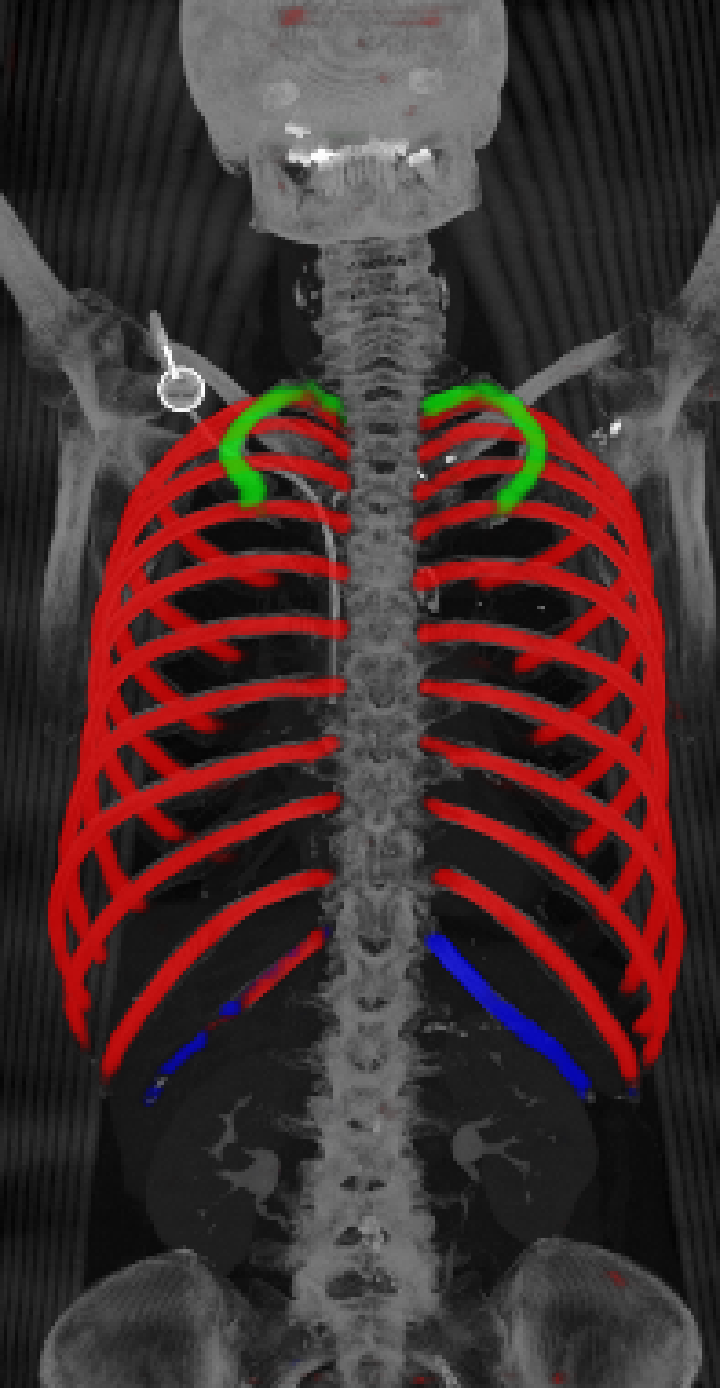}}
\subfigure[]{\includegraphics[height=60mm]{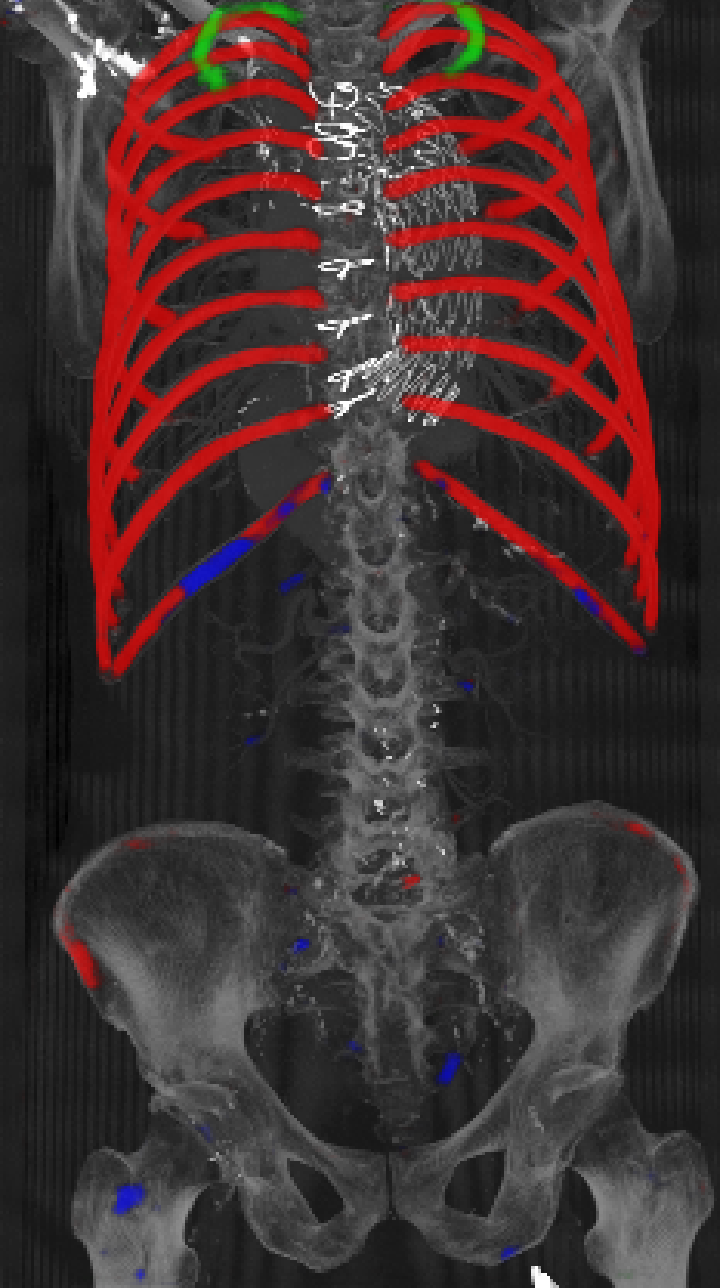}}
\subfigure[]{\includegraphics[height=62.2mm]{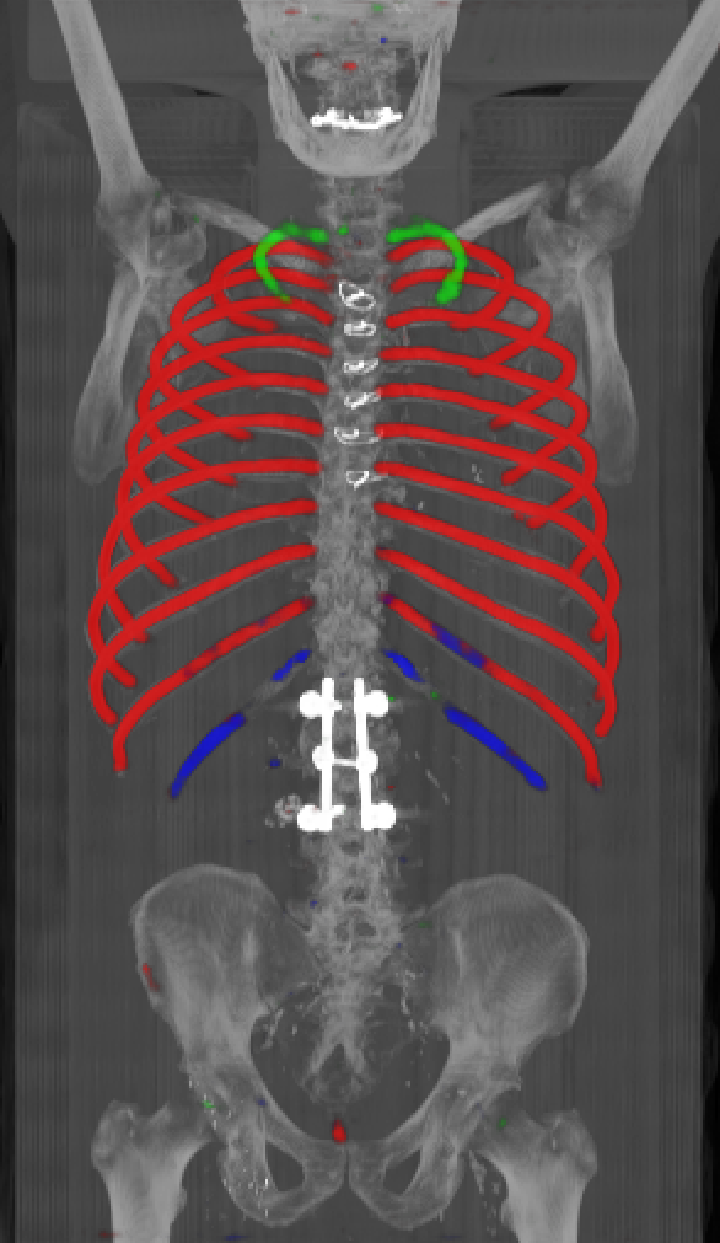}}
\subfigure[]{\includegraphics[height=62.2mm]{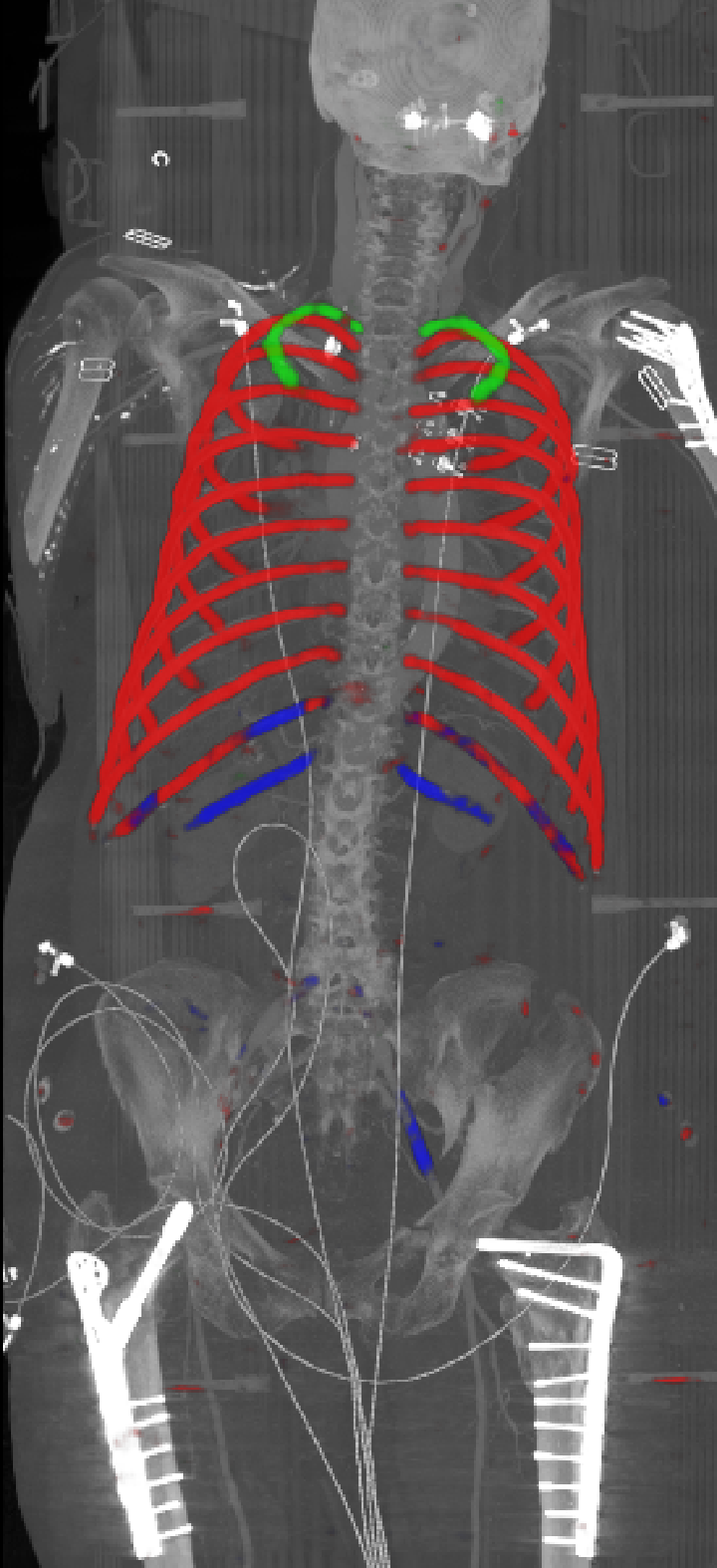}}
\subfigure[]{\includegraphics[height=62.2mm]{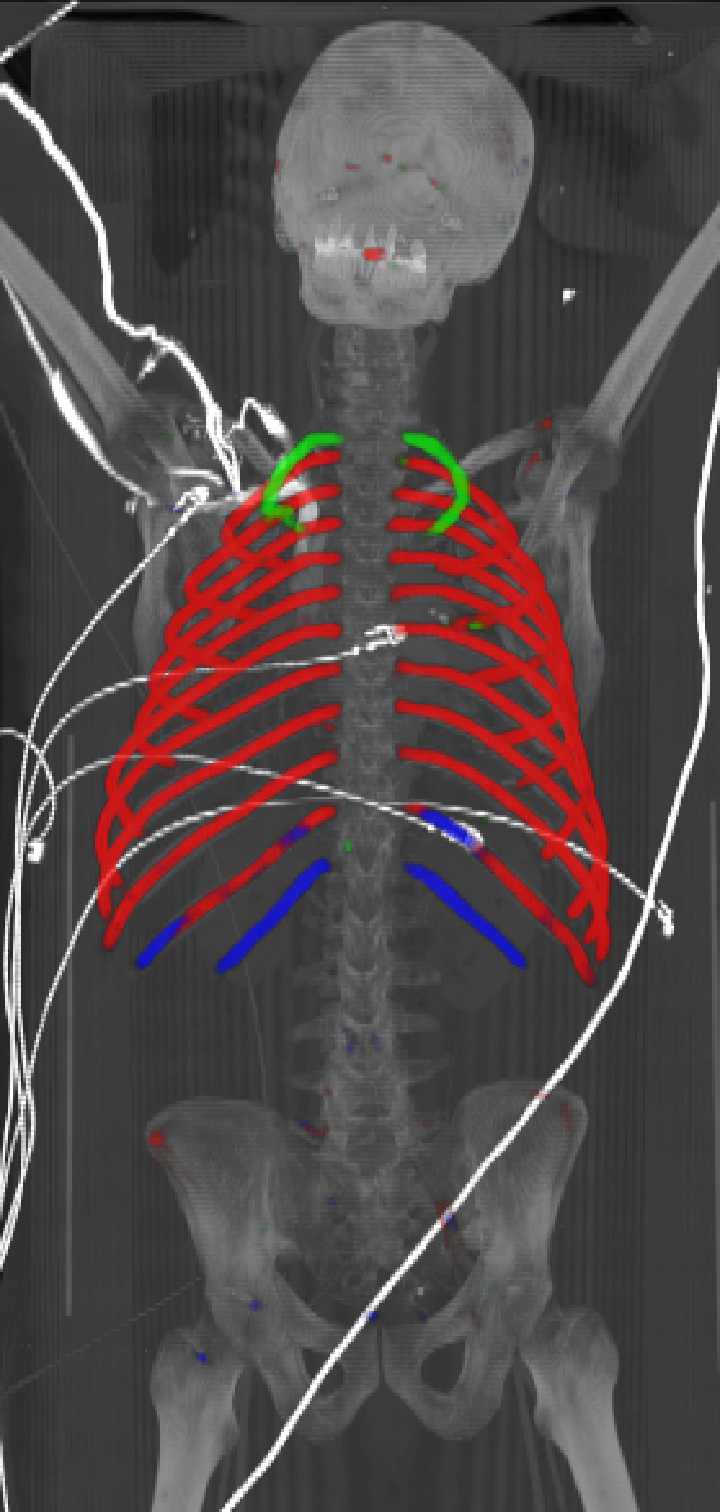}}
\subfigure[]{\includegraphics[height=62.2mm]{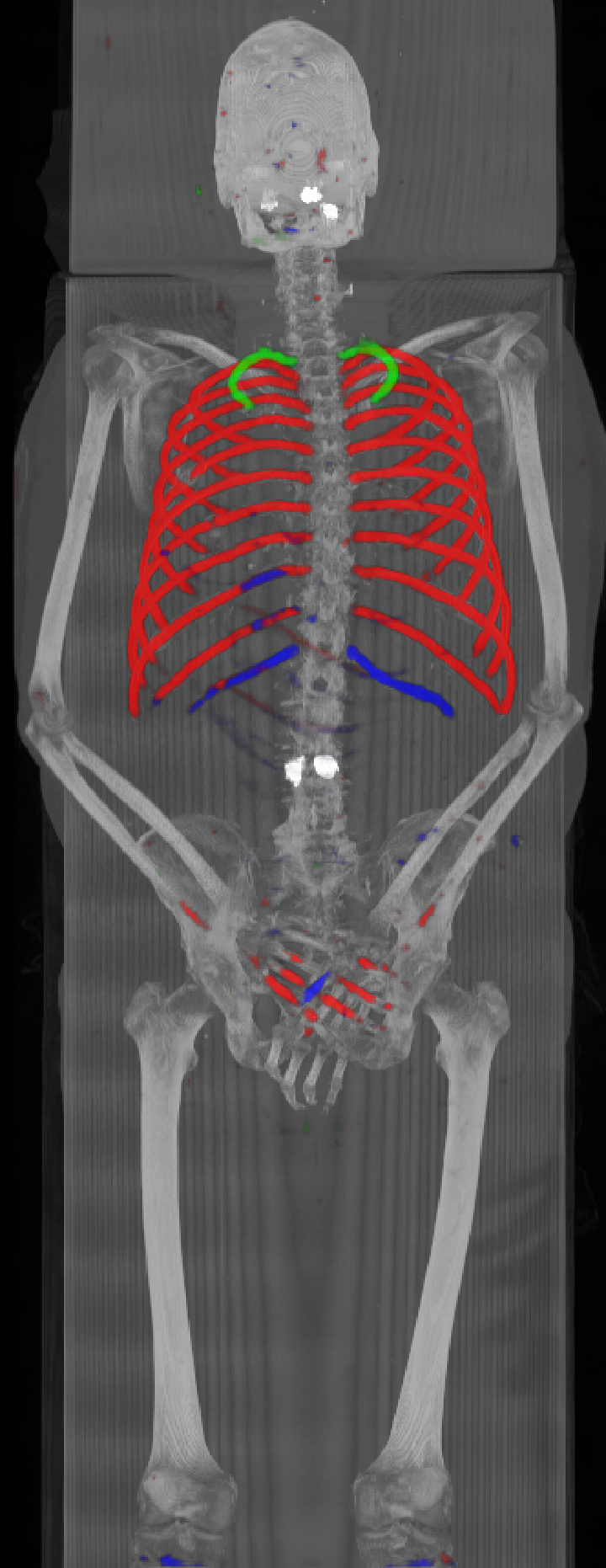}}
\subfigure[]{\includegraphics[height=26mm]{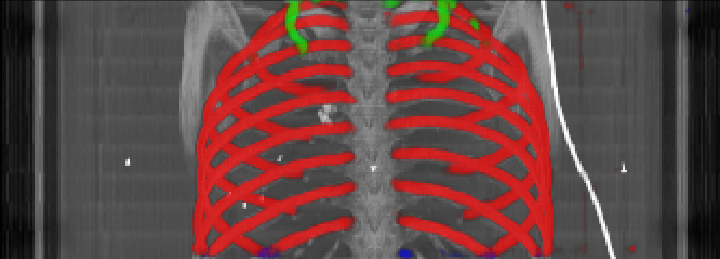}}
\subfigure[]{\includegraphics[height=26mm]{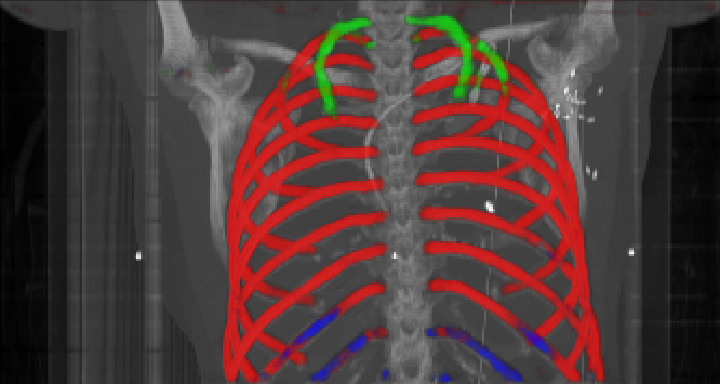}}
\caption{Maximum intensity projections (MIP) of selected CT volumes overlaid with the multi-label output of the neural network (green: first rib; red: intermediate rib; blue: twelfth rib). The selected case above display common difficulties which are inherent in the data set, such as pads (a) or cables (b), internal devices such as pacemakers (c), stents (d), spinal (e) and femural/humeral implants (f), injected contrast agents (g), patient shape variations such as scoliosis (h), limited field of views (FOVs), i.e. partly missing first (i) or twelfth rib (j).}
\label{MIPS_difficult3}
\end{figure}

\begin{figure}
\centering
\subfigure[]{\includegraphics[height=60mm]{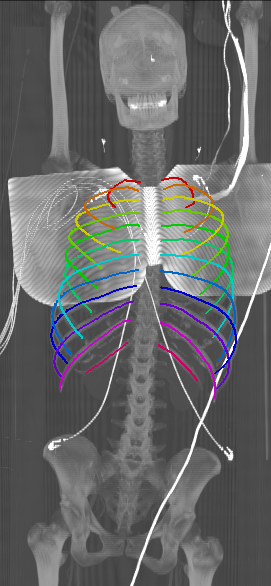}}
\subfigure[]{\includegraphics[height=60mm]{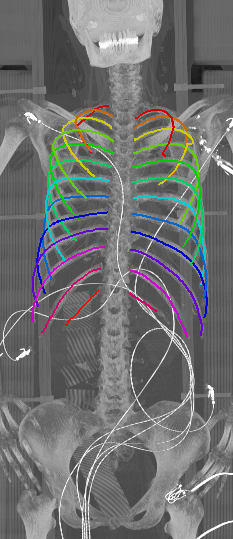}}
\subfigure[]{\includegraphics[height=60mm]{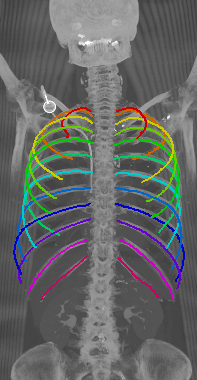}}
\subfigure[]{\includegraphics[height=60mm]{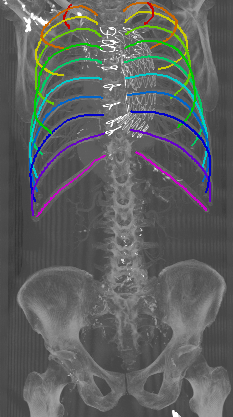}}
\subfigure[]{\includegraphics[height=62.2mm]{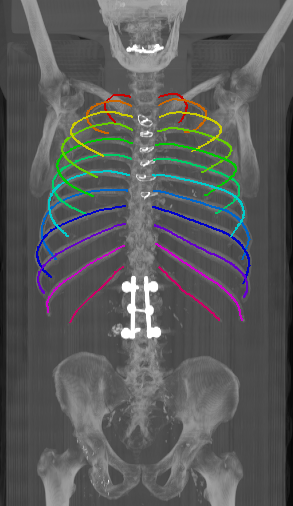}}
\subfigure[]{\includegraphics[height=62.2mm]{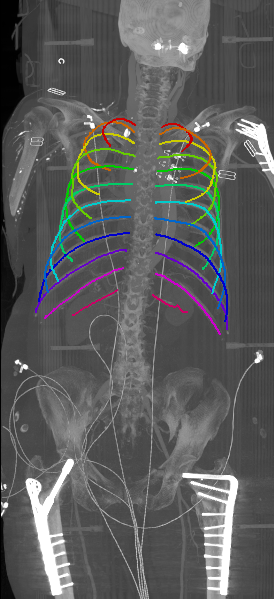}}
\subfigure[]{\includegraphics[height=62.2mm]{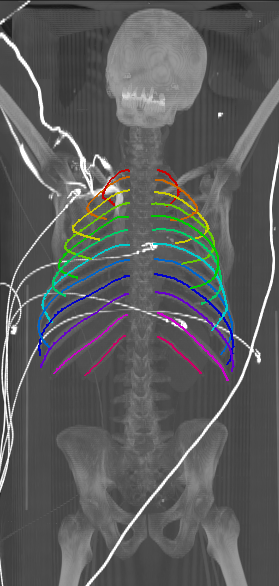}}
\subfigure[]{\includegraphics[height=62.2mm]{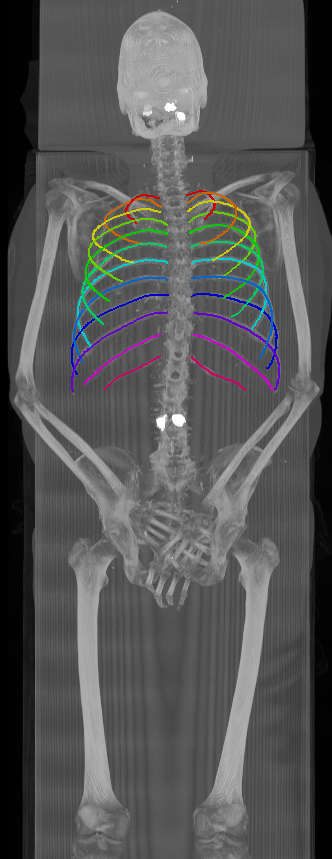}}
\subfigure[]{\includegraphics[height=26mm]{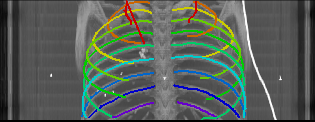}}
\subfigure[]{\includegraphics[height=26mm]{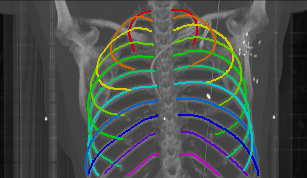}}
\caption{Automatically generated centerline splines associated with the FCNN outputs displayed in Figure \ref{MIPS_difficult3}. The selected case above display common difficulties which are inherent in the data set, such as pads (a) or cables (b), internal devices such as pacemakers (c), stents (d), spinal (e) and femural/humeral implants (f), injected contrast agents (g), patient shape variations such as scoliosis (h), limited field of views (FOVs), i.e. partly missing first (i) or twelfth rib (j).}
\label{Centerlines_difficult3}
\end{figure}

With an average Euclidean distance error of 0.787 mm, we obtained an overall result that is generally better compared to what is reported in the state of the art. Although, it needs to be kept in mind that results are unfortunately not directly comparable as both the data sets as well the evaluation metrics significantly differ across prior work. Similarly to the results obtained on the probability maps, distance errors are significantly higher for first and twelfth rib compared to the rest of the rib cage. As discussed, this is caused by the intrinsic challenges of these ribs, but certainly also an affect of error propagation in that sense that the quality of the probability maps also impacts centerline extraction. Interestingly, the right ribs are generally slightly worse compared to the left ribs, probably due to a slightly unbalanced data set with more challenges on the right side. Figure \ref{Centerlines_difficult3} shows the centerlines which were automatically generated using our walker algorithm from the corresponding network outputs displayed in Figure \ref{MIPS_difficult3}.

\section{Conclusion}  
We presented a fully automated two-stage approach for rib centerline extraction and labelling from CT images. 
First, multi-label probability maps (containing the classes first rib, twelfth rib, intermediate ribs, background) are calculated using a fully convolutional neural network and then centerlines are extracted from this multi-label information using a tracing algorithm.
For assessment,we performed a 4-fold cross validation on a set of 116 cases which includes several cases displaying typical clinical challenges. Comparing the automated extraction results to our manual ground truth, we were able to achieve an Euclidean distance error of 0.787 mm.
The 4-class label detection was crucial to simplify rib labelling by taking the network responses associated to the classes \emph{first rib} and \emph{twelfth rib} into account. Compared to a distinct detection of first and twelfth rib using separate networks, our multi-label task was chosen as it is memory and run-time efficient with negligible loss in final centerline accuracy.

In contrast to other approaches, no strong anatomical prior knowledge, e.g., in the form of geometrical models, was explicitly encoded into our pipeline to deal with pathological deviations.
Future work will focus on improving the performance of the neural network by using motion field and registration based data augmentation techniques and a more advanced data-driven image preprocessing. Moreover, we are currently investigating further improvements of our walker algorithm and the network architecture.


\begin{table}
\centering
	\begin{tabular}{c|ccccc}
		\
		\bf{rib} & \bf{sens.} & \bf{prec.} & \bf{dist.(mm)} & \bf{Dice}\\
		\hline
		01l & 0.878 & 0.927 & 1.431 & 0.902 \\
		01r & 0.871 & 0.915 & 1.511 & 0.894 \\
		02l & 0.972 & 0.971 & 0.878 & 0.972 \\
		02r & 0.952 & 0.963 & 0.880 & 0.957 \\
		03l & 0.981 & 0.984 & 0.782 & 0.983 \\
		03r & 0.976 & 0.973 & 0.854 & 0.975 \\
		04l & 0.989 & 0.994 & 0.740 & 0.992 \\
		04r & 0.984 & 0.984 & 0.785 & 0.984 \\
		05l & 0.994 & 0.993 & 0.763 & 0.993 \\
		05r & 0.972 & 0.978 & 0.811 & 0.975 \\
		06l & 0.994 & 0.992 & 0.757 & 0.993 \\
		06r & 0.964 & 0.965 & 0.757 & 0.964 \\
		07l & 0.991 & 0.995 & 0.731 & 0.993 \\
		07r & 0.969 & 0.957 & 0.721 & 0.963 \\
		08l & 0.992 & 0.984 & 0.730 & 0.988 \\
		08r & 0.969 & 0.968 & 0.719 & 0.969 \\
		09l & 0.993 & 0.993 & 0.738 & 0.993 \\
		09r & 0.973 & 0.971 & 0.699 & 0.972 \\
		10l & 0.986 & 0.992 & 0.715 & 0.989 \\
		10r & 0.965 & 0.969 & 0.674 & 0.967 \\
		11l & 0.984 & 0.974 & 0.732 & 0.979 \\
		11r & 0.961 & 0.962 & 0.665 & 0.962 \\
		12l & 0.917 & 0.972 & 0.921 & 0.944 \\
		12r & 0.886 & 0.942 & 0.905 & 0.914 \\
		\hline
		\bf{all ribs} & \bf{0.972} & \bf{0.976} & \bf{0.787} & \bf{0.974} \\
		\multicolumn{4}{c}{~~}\\
	\end{tabular}
	\caption{Rib-wise evaluation of the method based on the final labeled centerline point sets. A detected rib centerline point counts only as true positive if the correct label was determined. The table shows the summary for the collected 116 cases and reports sensitivity, precision, Euclidean distance and Dice score.\label{tbl_result_ctr_line_labeled}}
\end{table}

\begin{center}
	\begin{table}
		\centering
		\begin{tabular}{c|c}
			\
			\bf{No. missed ribs} & \bf{Case percentage} \\
			\hline
			0 & 81.6 \% \\
			1 & 8.8 \% \\
			2 & 7.0 \% \\
			$\geq$3 & 2.6 \% \\
			\multicolumn{2}{c}{~~}\\
		\end{tabular}
		\caption{Percentage of cases with missed labeled ribs. A rib counts as missed, if less than half of the ground truth rib centerline could be detected. A detected rib centerline point counts only as true positive if the correct label was determined.\label{tbl_result_ctr_line_casewise}}
	\end{table}
\end{center}

\end{document}